\documentclass[runningheads]{llncs}

\usepackage[utf8]{inputenc}
\usepackage[T1]{fontenc}
\usepackage[english]{babel}

\usepackage[table,xcdraw]{xcolor}

\usepackage{amsmath,amssymb,amsfonts}
\usepackage{textcomp}
\usepackage{cuted}
\usepackage{subcaption}

\usepackage{graphicx}
\usepackage{tikz}
\usepackage{amsmath}
\usepackage{booktabs}  
\usepackage{tabularx}  
\usepackage{multirow}  

\usepackage{algorithm}
\usepackage{algorithmic}
\usepackage{fontawesome5} 
\usepackage{caption}      
\usepackage{cite}
\usepackage{hyperref}
\usepackage{array}
\usepackage{threeparttable}

\definecolor{headerColor}{RGB}{52, 58, 64}   
\definecolor{groupBg}{RGB}{233, 236, 239}    
\definecolor{rowGray}{RGB}{248, 249, 250}    
\definecolor{tagCyan}{HTML}{17A2B8}
\definecolor{tagBlue}{HTML}{007BFF}
\definecolor{tagPurple}{HTML}{6F42C1}
\definecolor{tagGreen}{HTML}{28A745}
\definecolor{tagOrange}{HTML}{FD7E14}
\definecolor{tagRed}{HTML}{DC3545}
\definecolor{tagGray}{HTML}{6C757D} 
\definecolor{dividerGray}{RGB}{220, 220, 220}

\newcolumntype{L}{>{\raggedright\arraybackslash}X}

\usetikzlibrary{shapes.geometric, arrows.meta, positioning, fit, backgrounds, calc, shadows}
\def\BibTeX{{\rm B\kern-.05em{\sc i\kern-.025em b}\kern-.08em
		T\kern-.1667em\lower.7ex\hbox{E}\kern-.125emX}}


\begin{document}


\title{GloResNet: A lightweight 3D CNN with global topological features for preterm brain injury prediction}

\author{Boyu Yuan\inst{1} \and
	Jiamiao Lu\inst{1} \and
	Weichuan Zhang\inst{1} \and
	Benqing Wu\inst{2} \and
	Tuo Wang\inst{3} \and
	Changshan Wang\inst{2} \and
	Changming Sun\inst{4} \and
	Liang Guo\inst{2}}
\authorrunning{B. Yuan et al.}
\institute{Image Computing Laboratory, Shaanxi University of Science and Technology, Xi'an, China\\
	\email{202315030421@sust.edu.cn} (B.Y.), \email{241612058@sust.edu.cn} (J.L.), \email{zwc2003@163.com} (W.Z.) \and
	Department of Neonatology, Shenzhen University of Advanced Technology General Hospital, Shenzhen, China\\
	\email{wubenqing783@126.com} (B.W.), \email{cs.wang1@siat.ac.cn} (C.W.), \email{419221185@qq.com} (L.G.) \and
	Department of Neurosurgery, The First Affiliated Hospital of Xi'an Jiaotong University, Xi'an, China\\
	\email{wt1972@tom.com} (T.W.) \and
	CSIRO Technology, PO Box 76, Epping, NSW 1710, Australia\\
	\email{changming.sun@csiro.au} (C.S.)}

\maketitle

\begin{abstract}
	\noindent 
This study introduces an automated deep learning framework for predicting brain injury (BI) in preterm infants from T2-weighted MRI (dHCP dataset). We propose GloResNet, a lightweight 3D CNN based on ResNet-10, pretrained on MedicalNet to address data scarcity. A global manifold mapping strategy first resamples each 3D volume to $128\times128\times128$ and then applies subject-wise $z$-score intensity normalization, thereby preserving global topology while standardizing appearance. Training integrates mixup, class weighting, and test-time augmentation for robustness. In 5-fold cross-validation, GloResNet achieved 75.18\% average accuracy (peak 81.82\%), with specificity 0.81 and sensitivity 0.76. Results demonstrate that a topology-aware lightweight CNN has the capability to effectively predict neonatal BI, offering a non-invasive screening tool. The source code of this paper can be obtained from the GitHub repository: https://github.com/ICL-SUST/GloResNet-Preterm-Brain

\end{abstract}

	\section{Introduction}\label{sec:introduction}
	
Preterm birth, affecting about 15 million neonates annually, is a major cause of morbidity and mortality. The cerebrovascular and neural immaturity in these infants makes them highly vulnerable to brain injuries (BI) such as periventricular leukomalacia, intraventricular hemorrhage, and diffuse white matter injury. These early injuries are strongly linked to adverse long-term neurodevelopmental outcomes, including cerebral palsy and cognitive deficits~\cite{10.1093/brain/awab118,10.1542/peds.2022-057442,georgePPREMOProspectiveCohort2015,andersonPredictingNeurodevelopmentalOutcome2023a}. While T2-weighted MRI is the gold standard for diagnosing neonatal BI, its clinical interpretation relies heavily on radiologists' subjective judgment, leading to labor-intensive processes and inter-observer variability \cite{10.3389/fped.2022.961556,auto_ref_20,app13053211}. The availability of high-resolution datasets such as the Developing Human Connectome Project (dHCP)~\cite{auto_ref_4,10.1162/imag_a_00512} now presents a crucial opportunity to develop objective, automated early screening tools using deep learning~\cite{song2025efficient, lu2026meningioma, guo2026gattenrnn, ren2026deep, liao2026neuron, liao2025learning,  zhang2026adaptive, wang2026dual}.

Existing image feature extraction techniques~\cite{jing2022recent,  jing2022image, liu2024aekan, zhang2024re,  qiu2021recurrent,  pan2024pseudo, ren2024few, jing2023ecfrnet,  pan2025overcoming,  lei2024semi, pan2024dycr} have shifted from manual algorithms~\cite{shui2013corner, zhang2014corner, zhang2019corner, shui2012noise, zhang2017noise, zhang2015contour, zhang2020corner, gao2020fast, li2019multi, zhang2019discrete,  wang2020corner, li2023traffic, zhang2023image,  xie2026second, bao2022corner, an2023edge, li2023mutual} to deep learning architectures~\cite{ zhang2021ndpnet, ma2023ct, liao2025dynamic,  li2023m, wang2024unbiased, lu2022image, jing2021novel, zheng2023fully,  ren2025adaptive, wang2025principal, liao2022asrsnet, li2024automotive,  liao2023feature, tang2025cascading, ren2025zero, wang2025feature} for feature extraction. Although convolutional neural networks (CNNs) have advanced medical image analysis~\cite{auto_ref_18,auto_ref_1,bioengineering10121435}, automated prediction of preterm BI remains a significant challenge. This is largely due to the tension between high-dimensional 3D MRI data and the scarcity of annotated preterm infant samples, which often number only in the hundreds~\cite{10.3389/fninf.2022.1006532,10977025}. To manage this, many existing methods use a divide-and-conquer approach, processing slices or patches. However, while reducing computational cost and preserving local details, this strategy fragments the brain's global topology~\cite{auto_ref_5,11370476}. Since key preterm BI pathologies—such as ventriculomegaly or tissue atrophy—are inherently global, such localized analysis often fails to capture essential macroscopic contextual information, limiting the model's diagnostic capability for structural injuries~\cite{sym17071108,10778607,10130343}.

To address the challenge of automated BI prediction in preterm infants with limited data, this paper proposes GloResNet, a lightweight 3D CNN framework. Departing from deep, high-resolution models, we adopt a ``Less is More'' principle via a global manifold mapping strategy that resamples 3D MRI volumes to $128\times128\times128$ and then applies subject-wise $z$-score normalization~\cite{11063450,10083150}, preserving global brain topology to capture macroscopic injury features. To prevent overfitting, a ResNet-10 backbone is initialized with MedicalNet pretrained weights and enhanced with mixup and test-time augmentation~\cite{auto_ref_3,auto_ref_22,math12244003}. Evaluated on the dHCP dataset using 5-fold cross-validation, GloResNet achieved an average accuracy of 75.18\% (peak 81.82\%), with balanced sensitivity (0.76) and specificity (0.81). Results confirm that a lightweight, topology-aware model offers strong clinical applicability in data-scarce medical imaging contexts~\cite{fi17120562,auto_ref_16,10223434}.

For clarity, the ``global manifold mapping'' used in the remainder of the paper is the composition of trilinear resampling to $128\times128\times128$ and subject-wise $z$-score intensity normalization, rather than geometric resizing alone.

	\section{Related Work}
	
	\subsection{Deep Learning in Neonatal Brain MRI Analysis}

The early diagnosis of neonatal BI initially utilized radiomics and traditional machine learning, though these were constrained by the subjectivity of manual feature engineering. The rise of deep learning has established CNNs as the primary approach. While 2D slice-based CNNs offer computational efficiency, they overlook inter-slice relationships and lack 3D contextual information critical for detecting diffuse injuries. In comparison, 3D CNNs process volumetric data directly and excel at capturing spatial features, but their high parameter counts increase overfitting risks, especially in the small-sample settings common in preterm infant research.

	\subsection{Local Patches vs. Global Context}
	
To tackle the computational demands of high‑dimensional 3D MRI, patch‑based strategies have been widely adopted. While preserving fine textural details, these methods disrupt the global topological connections of brain anatomy. Preterm BIs such as Periventricular Leukomalacia (PVL) often involve macroscopic structural changes—such as ventriculomegaly or overall volume loss—rather than localized texture variations. A limited field of view therefore restricts the detection of such pathological deformations. As a result, recent approaches have shifted toward global methods that retain full anatomical context through down‑sampling, proving more effective for identifying macroscopic injuries.
	
	\subsection{Transfer Learning and Data Scarcity}
	
Data scarcity severely limits medical image analysis~\cite{11095799}. Training deep 3D networks from scratch on small datasets such as dHCP (a few hundred subjects) leads to poor generalization. Transfer learning offers a solution~\cite{10977025,ouyangMachinelearningBasedPrediction2024a}. While a common approach inflates 2D ImageNet weights to 3D, the large domain gap with medical images reduces its effectiveness~\cite{auto_ref_21}. In contrast, models such as MedicalNet, pre-trained on large-scale medical data, provide stronger initialization for 3D anatomical features~\cite{awudongAttentionalAdversarialTraining2024}. This work employs such domain-specific transfer learning combined with a lightweight network design to efficiently extract 3D features under data-scarce conditions.

	\section{Methodology}
	
	\begin{figure}[!htbp]
		\includegraphics[width=1\textwidth]{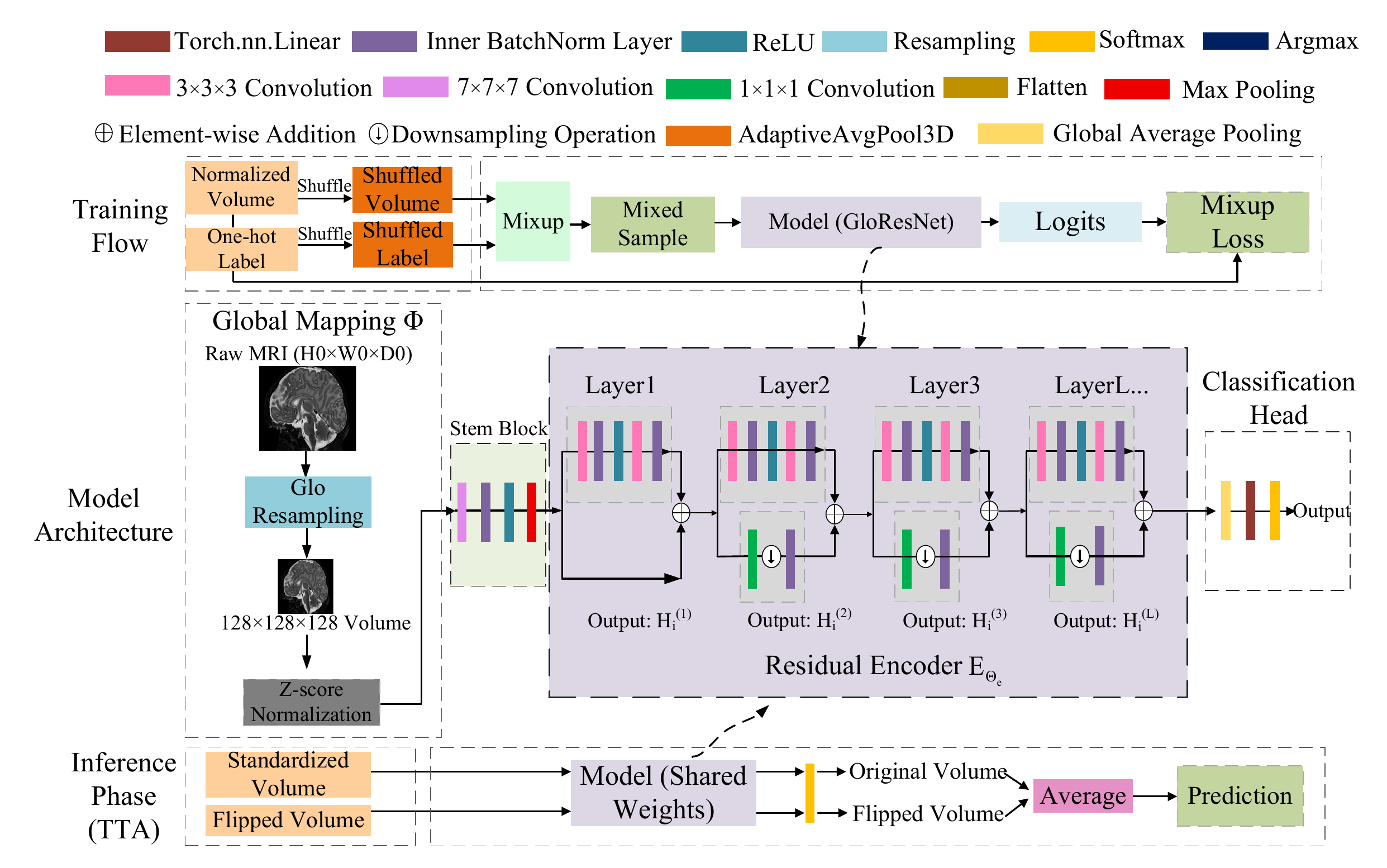}
		\caption{The proposed GloResNet framework follows a three-stage pipeline: (1) Global Mapping ($\Phi$), which resamples each high-resolution MRI scan to a unified $128 \times 128 \times 128$ space and then applies subject-wise $z$-score intensity normalization to preserve anatomical topology while standardizing appearance; Residual Feature Encoding, which employs a MedicalNet-initialized ResNet-10 backbone as the Residual Encoder ($E_{\Theta_c}$) to efficiently extract discriminative 3D representations with minimal parameters; and (3) Regularized Optimization, implemented via mixup augmentation during the Training Flow and test-time averaging (TTA) during the Inference Phase to refine decision boundaries and improve generalization on limited data.}
		\label{fig:step_workflow}
	\end{figure}
	
	\subsection{Problem Formulation and GloResNet Framework}

This work formulates automated preterm BI screening as a supervised binary classification problem on volumetric MRI. Let $\mathcal{V}_{\mathrm{raw}} \subset \mathbb{R}^{H_0 \times W_0 \times D_0}$ denote the space of raw T2-weighted scans, let $\mathcal{Y}=\{0,1\}$ denote the label space, and let the training set be $\mathcal{D}=\{(\mathbf{V}^{\mathrm{raw}}_i,z_i)\}_{i=1}^{n}$, where $i$ is the sample index, $n$ denotes the total number of training samples, $z_i=0$ indicates normal development, and $z_i=1$ indicates BI. To keep the derivation unambiguous, we reserve distinct symbols for each stage of the pipeline: $\mathbf{V}^{\mathrm{raw}}_i$ is the raw scan, $\mathbf{V}^{\mathrm{res}}_i$ is the resampled scan, $\mathbf{V}_i$ is the intensity-normalized scan, $\mathbf{h}_i$ is the latent representation, and $\mathbf{p}_i$ is the predicted class-probability vector.

To tackle data scarcity ($n \ll H_0W_0D_0$), GloResNet is organized as a senquential pipeline of global perception, lightweight encoding, and regularized optimization (Fig.~\ref{fig:step_workflow}). The full forward path is
\begin{equation}
	\begin{aligned}
		\mathbf{V}^{\mathrm{res}}_i &= R(\mathbf{V}^{\mathrm{raw}}_i; S), \\
		\mathbf{V}_i &= \Phi(\mathbf{V}^{\mathrm{raw}}_i; S) := \mathcal{N}(\mathbf{V}^{\mathrm{res}}_i), \\
		\mathbf{h}_i &= E_{\Theta_e}(\mathbf{V}_i), \qquad
		\mathbf{p}_i = q_{\Theta}(\mathbf{V}_i) := C_{\Theta_c}(\mathbf{h}_i),
	\end{aligned}
	\label{eq:forward}
\end{equation}
where $S=128$ is the target spatial size, $R$ denotes geometric resampling, $\mathcal{N}$ denotes intensity normalization, $\Phi = \mathcal{N} \circ R$ is the complete global manifold mapping operator (with $\circ$ denoting function composition: $(\mathcal{N} \circ R)(\mathbf{V}^{\mathrm{raw}}_i) = \mathcal{N}(R(\mathbf{V}^{\mathrm{raw}}_i))$, i.e., resampling first, normalization second), $E_{\Theta_e}$ is the 3D encoder, $C_{\Theta_c}$ is the classification head, and $\Theta=\{\Theta_e,\Theta_c\}$ collects all trainable parameters. Accordingly, model training seeks
\begin{equation}
	\Theta^{\star} = \arg\min_{\Theta} \mathbb{E}_{(\mathbf{V}^{\mathrm{raw}},z)\sim \mathcal{D}}
	\left[\ell\!\left(q_{\Theta}\!\left(\Phi(\mathbf{V}^{\mathrm{raw}}; S)\right), z\right)\right].
	\label{eq:objective}
\end{equation}

	\subsection{Global Topological Perception and 3D Feature Encoding}
	
	1. \textbf{Motivation}: From Local Patches to Global Topology \cite{auto_ref_19,11370476}
	
	Traditional 3D image analysis often employs patch-based strategies to mitigate memory constraints. However, preterm BI is frequently manifested as macroscopic deformations of anatomical structures, such as ventriculomegaly caused by white matter softening or a reduction in parenchymal volume. Patching operations disrupt the global spatial continuity of brain tissues, causing the model to lose critical relative positional information.
	
	To preserve holistic anatomical relationships, we introduce a global manifold mapping strategy that transforms each raw scan $\mathbf{V}^{\mathrm{raw}}_i$ into a standardized volume $\mathbf{V}_i \in \mathbb{R}^{S \times S \times S}$ with $S=128$. The mapping is decomposed into geometric resampling followed by intensity normalization, as defined in Eq.~\eqref{eq:forward}. Let $\Omega_S = \{1,\dots,S\}^3$ be the target lattice and $a \in \{1,2,3\}$ index the three spatial dimensions. For any target coordinate $\mathbf{x} \in \Omega_S$, the resampled intensity is computed by trilinear interpolation over the eight nearest voxels:
	\begin{equation}
		\mathbf{V}^{\mathrm{res}}_i(\mathbf{x}) =
		\sum_{\mathbf{r} \in \mathcal{R}(\Psi_i(\mathbf{x}))}
		\mathbf{V}^{\mathrm{raw}}_i(\mathbf{r})
		\prod_{a=1}^{3}\left(1-\left|\Psi_i(\mathbf{x})_a - r_a\right|\right),
		\label{eq:resample}
	\end{equation}
	where $\Psi_i:\Omega_S \rightarrow \mathbb{R}^{3}$ maps a target coordinate to its continuous coordinate in the original scan, $\mathbf{r}$ denotes the integer coordinates of a voxel in the original scan, and $\mathcal{R}(\Psi_i(\mathbf{x}))$ is the set of the $2\times2\times2$ voxel indices surrounding $\Psi_i(\mathbf{x})$. We then apply subject-wise $z$-score normalization:
	\begin{equation}
		\mathbf{V}_i =
		\mathcal{N}(\mathbf{V}^{\mathrm{res}}_i)
		=
		\frac{\mathbf{V}^{\mathrm{res}}_i - \mu_i}{\sigma_i + \varepsilon},
		\label{eq:normalize}
	\end{equation}
	where $\mu_i$ and $\sigma_i$ are the mean and standard deviation of all voxels in $\mathbf{V}^{\mathrm{res}}_i$, and $\varepsilon>0$ is a small constant for numerical stability. Together, Eqs.~\eqref{eq:resample} and \eqref{eq:normalize} fully specify the operator $\Phi = \mathcal{N} \circ R$.
	
	2. \textbf{Architecture}: The ``Less is More'' 3D Backbone \cite{10223434,wuSDSNetLightweight3D2023}
	
	With global inputs secured, the core challenge shifts to designing an effective feature extractor. Given that the parameters of 3D convolutions grow cubically with kernel size, and the dHCP cohort is limited in size, blindly employing deep networks (e.g., ResNet-50) inevitably leads to overfitting. Adhering to the ``Less is More'' principle, we construct a lightweight GloResNet-10.
	
	The backbone comprises an initial convolutional stem and four residual stages, with channel schedule $\boldsymbol{\chi} = [64, 128, 256, 512]$. Let $l$ denote the stage index, and let $\mathbf{H}^{(l-1)}$ and $\mathbf{H}^{(l)}$ denote the input and output feature tensors of stage $l$. Let $C_{l}$ be the number of output channels at stage $l$ (so $C_1=64$, $C_2=128$, $C_3=256$, $C_4=512$, and $C_0=1$ for the input). For a spatial position $\mathbf{u}$ in the feature map, a standard 3D convolution is written as
	\begin{equation}
		\mathbf{H}_q^{(l)}(\mathbf{u}) =
		\sigma\!\left(
		b_q^{(l)} +
		\sum_{j=1}^{C_{l-1}}
		\sum_{\boldsymbol{\delta}\in\Delta}
		K_{q,j}^{(l)}(\boldsymbol{\delta})\,
		\mathbf{H}_j^{(l-1)}(\mathbf{u}+\boldsymbol{\delta})
		\right),
		\label{eq:conv3d}
	\end{equation}
	where $\Delta=\{-1,0,1\}^3$ is the $3\times3\times3$ kernel support, $j=1,\dots,C_{l-1}$ indexes the input channels, $q=1,\dots,C_{l}$ indexes the output channels, and $\sigma(\cdot)$ is the ReLU activation function. To mitigate gradient degradation, each block then adds a shortcut branch to the transformed features:
	\begin{equation}
		\mathbf{H}^{(l)}_{\mathrm{res}} =
		\mathcal{R}^{(l)}\!\left(\mathbf{H}^{(l-1)}; \Theta_r^{(l)}\right) +
		\mathcal{S}^{(l)}\!\left(\mathbf{H}^{(l-1)}\right).
		\label{eq:residual}
	\end{equation}
	Here, $\mathcal{R}^{(l)}$ denotes the residual transform and $\mathcal{S}^{(l)}$ denotes the shortcut mapping. Let $L=4$ be the total number of stages, and let $\mathbf{H}^{(L)}_i$ denote the output tensor of the final residual stage for subject $i$. The encoder-to-classifier interface, where $\mathrm{GAP}$ denotes global average pooling, is then
	\begin{equation}
		\mathbf{h}_i = \mathrm{GAP}\!\left(\mathbf{H}^{(L)}_i\right), \qquad
		\mathbf{p}_i = \mathrm{softmax}\!\left(\mathbf{W}_{\mathrm{cls}}\mathbf{h}_i + \mathbf{b}_{\mathrm{cls}}\right),
		\label{eq:head}
	\end{equation}
	where $\mathbf{h}_i$ is the latent feature vector, $\mathbf{p}_i$ is the predicted class probability vector, and $\mathbf{W}_{\mathrm{cls}},\mathbf{b}_{\mathrm{cls}}$ are the weight matrix and bias of the linear classification layer. This explicitly closes the mapping from the stage-level features in Eqs.~\eqref{eq:conv3d} and \eqref{eq:residual} to the prediction rule in Eq.~\eqref{eq:forward}.
	
	3. \textbf{Initialization}: Domain Knowledge Transfer via MedicalNet
	
	Despite the lightweight nature of ResNet-10, training from scratch on small samples remains difficult due to a lack of inductive bias. To solve this ``cold start'' problem, we employ transfer learning \cite{9802503}. Specifically, the encoder parameters are initialized from MedicalNet, while the classifier head is initialized independently. Let $\Theta_e^{(0)}$ and $\Theta_c^{(0)}$ denote the initial parameter values before training. Then:
	\begin{equation}
		\Theta_e^{(0)} \leftarrow \Theta_e^{\mathrm{med}}, \qquad
		\Theta_c^{(0)} \sim \text{Gaussian}(0, \sigma_0^2 \mathbf{I}),
		\label{eq:init}
	\end{equation}
	where $\Theta_e^{\mathrm{med}}$ denotes the pretrained MedicalNet weights, $\mathbf{I}$ is the identity matrix, and we set $\sigma_0 = 0.01$. This step follows naturally from Eq.~\eqref{eq:forward}, because only the encoder $E_{\Theta_e}$ receives transferable anatomical priors.
	
	\subsection{Regularized Optimization and Inference Strategy}
	
	1. \textbf{Theoretical Basis}: Variational Risk Minimization (VRM)
	
	Under limited data, empirical risk minimization (ERM) encourages memorization of training samples and produces sharp decision boundaries. We therefore introduce a variational risk minimization (VRM) technique using mixup for promoting smoother decision regions.
	
	Based on the assumption that the data manifold is locally linear, mixup is applied to the standardized volumes $\mathbf{V}_i$ produced by Eq.~\eqref{eq:normalize}, rather than to the raw scans. Let $\rho$ be a random permutation over a mini-batch, let $\mathbf{t}_i \in \{0,1\}^2$ denote the one-hot encoding of label $z_i$, and let $\eta_i \sim \mathrm{Beta}(\alpha,\alpha)$ with $\alpha=1.0$ be the sample-wise mixing coefficient. The mixed sample is then defined as
	\begin{equation}
		\begin{aligned}
			\widetilde{\mathbf{V}}_i &= \eta_i \mathbf{V}_i + (1 - \eta_i)\mathbf{V}_{\rho(i)}, \\
			\widetilde{\mathbf{t}}_i &= \eta_i \mathbf{t}_i + (1 - \eta_i)\mathbf{t}_{\rho(i)}.
		\end{aligned}
		\label{eq:mixup}
	\end{equation}
	This construction encourages approximately linear behavior of the predictor $q_{\Theta}$ between nearby samples:
	\begin{equation}
		q_{\Theta}(\widetilde{\mathbf{V}}_i)
		\approx
		\eta_i q_{\Theta}(\mathbf{V}_i) + (1 - \eta_i) q_{\Theta}(\mathbf{V}_{\rho(i)}),
		\label{eq:mixup_linear}
	\end{equation}
	which effectively regularizes the model by encouraging smooth transitions between samples, thereby improving generalization and robustness to input perturbations.
	
	2. \textbf{Loss Function}: Unifying Class Balance and Regularization \cite{luCollaborativeMultiMetadataFusion2023}
	
	Addressing the class imbalance problem where normal controls far outnumber injury cases, pure mixup might be dominated by the majority class \cite{auto_ref_20}. Under the mixup-induced vicinal distribution and class-balanced weighting, the generic loss in Eq.~\eqref{eq:objective} is instantiated as a weighted soft-label cross-entropy objective on the mixed targets in Eq.~\eqref{eq:mixup}. Let $n$ be the total number of training samples in the current fold, and let $n_c$ be the number of samples of class $c$ ($c=0$ for normal, $c=1$ for injury). The inverse class frequency weight for class $c$ is defined as $\omega_c = \frac{n}{2 n_c}$, where the factor $2$ accounts for binary classification. For a mini-batch of size $B$, the training loss becomes
	\begin{equation}
		\mathcal{L}_{\mathrm{mix}}(\Theta) =
		-\frac{1}{B}
		\sum_{i=1}^{B}
		\sum_{c=0}^{1}
		\omega_c\, \widetilde{t}_{i,c}\,
		\log \big(q_{\Theta}(\widetilde{\mathbf{V}}_i)_c\big).
		\label{eq:mixloss}
	\end{equation}
	
	Mathematically, this formulation imposes dual constraints: $\omega_c$ increases the penalty on the minority class, while the soft target $\widetilde{\mathbf{t}}_i$ prevents over-confident fitting and is therefore a direct consequence of the mixup construction in Eq.~\eqref{eq:mixup}.
	
	3. \textbf{Inference}: Test-Time Averaging (TTA) \cite{auto_ref_22,jimaging8050141,math12244003,moreiraAutomatedInfieldGrapevine2023}
	
	Upon completion of training, to further reduce prediction variance and leverage the bilateral symmetry of brain anatomy, we introduce test-time augmentation (TTA). For a raw input volume $\mathbf{V}^{\mathrm{raw}}_i$, the final probability vector is computed as
	\begin{equation}
		\bar{\mathbf{p}}(\mathbf{V}^{\mathrm{raw}}_i) =
		\frac{1}{2}
		\left[
		q_{\Theta}\!\left(\Phi(\mathbf{V}^{\mathrm{raw}}_i; S)\right) +
		q_{\Theta}\!\left(\mathcal{F}_{\text{lr}}\!\left(\Phi(\mathbf{V}^{\mathrm{raw}}_i; S)\right)\right)
		\right],
		\label{eq:tta}
	\end{equation}
	where $\mathcal{F}_{\text{lr}}$ denotes horizontal flipping along the left-right axis in the standardized space. The BI score used for decision making is the second component $\bar{\mathbf{p}}(\mathbf{V}^{\mathrm{raw}}_i)_1$ (index 1, corresponding to the injury class). This final equation closes the same raw-to-standardized entry point defined in Eq.~\eqref{eq:forward}.
	
	\section{Experiments}
	
	\subsection{Dataset Curation and Experimental Setup}
	\textbf{Dataset Description and Preprocessing} This study was conducted using T2-weighted (T2w) MRI scans from the dHCP dataset \cite{auto_ref_4,10.1162/imag_a_00512,9774943,Chiu2026.01.23.701337,Cullen2022.08.11.22278677,doi:10.1073/pnas.2121748119,fenchelNeonatalMultimodalCortical2022,maDevelopingHumanConnectome2025,Wu2021.06.24.449666,9774943}. T2w imaging is widely adopted in neonatal neuroimaging as it provides superior tissue contrast for the unmyelinated newborn brain, making it highly sensitive to subtle white matter abnormalities and structural alterations associated with preterm brain injury. The cohort comprises 128 subjects, categorized into a normal group ($N=74$) and an injury group ($N=54$) based on radiological scores. As detailed in Table \ref{tab:demographics}, there were no statistically significant differences ($P > 0.05$) between the two groups regarding gestational age at birth, birth weight, post-menstrual age (PMA) at scan, or biological sex. This ensures that the model's predictive performance is driven by BI features rather than confounding demographic variables.
	
	\begin{table}[htbp]
		\caption{Demographic characteristics of the study population. Values are presented as mean $\pm$ standard deviation or number (\%). $P$-values were calculated using independent two-sample $t$-tests for continuous variables and Pearson's $\chi^2$ test for categorical variables.}
		\label{tab:demographics}
		\centering
		\resizebox{\columnwidth}{!}{
			\begin{tabular}{lccc}
				\toprule
				\textbf{Characteristic} & \textbf{Normal Group} ($n=74$) & \textbf{Injury Group} ($n=54$) & \textbf{\textit{P}-value} \\
				\midrule
				Gestational age (wk) & 32.91 $\pm$ 2.98 & 31.96 $\pm$ 3.74 & 0.127 \\ 
				Birth weight (kg)    & 1.92 $\pm$ 0.66  & 1.74 $\pm$ 0.74  & 0.159 \\ 
				PMA at scan (wk)     & 36.35 $\pm$ 3.22 & 36.38 $\pm$ 3.75 & 0.961 \\ 
				Male, $n$ (\%)        & 48 (64.9\%)      & 30 (55.6\%)      & 0.377 \\ 
				\bottomrule
			\end{tabular}
		}
	\end{table}

	\noindent We selected T2-weighted sequences due to their higher contrast-to-noise ratio for visualizing unmyelinated white matter in preterm infants compared to T1-weighted scans. To align with the global manifold mapping defined in Eqs.~\eqref{eq:resample} and \eqref{eq:normalize}, all volumes underwent standardized preprocessing: skull-stripping and bias-field correction via the dHCP minimal pipeline, spatial normalization to $128 \times 128 \times 128$ voxels using trilinear interpolation, and per-subject intensity normalization via $z$-score standardization.

A stratified 5-fold cross-validation scheme~\cite{jangPredicting2yearNeurodevelopmental2024} was applied, with partitioning at the subject level to prevent data leakage. This ensures scans from the same subject do not appear in both training and validation sets, supporting a rigorous evaluation of the model’s generalization to unseen patients.

	\subsection{Implementation and Evaluation Protocols}
	\textbf{Implementation Details}. The GloResNet framework was implemented using the PyTorch library and trained on a workstation equipped with an NVIDIA RTX 2080Ti GPU (22GB VRAM). The backbone network (ResNet-10) was initialized with weights transfer-learned from MedicalNet to ensure robust feature extraction.
	
	\textbf{Hyperparameter Configuration}. Optimization was performed using the AdamW algorithm, which decouples weight decay from gradient updates, providing superior regularization compared to standard SGD. The training process spanned 150 epochs with a batch size of 32. We adopted a cosine annealing learning rate scheduler, initializing the learning rate at $2 \times 10^{-3}$ and decaying it to a minimum of $1 \times 10^{-6}$, facilitating convergence into flatter minima. To further enhance robustness, online data augmentation was applied, including random 3D rotations ($ \pm 10^{\circ}$), flipping, and the mixup strategy ($\alpha=1.0$) with a probability of 0.5.
	
	\textbf{Performance metrics}. The diagnostic performance was comprehensively evaluated using five quantitative metrics \cite{10879789}. 
	\begin{itemize}
		\item 
		uracy (Acc) and F1 provide a holistic view of classification correctness.
		\item Sensitivity (Sens) measures the ability to correctly identify injury cases, which is the primary safety metric for clinical screening.
		\item Area under the curve (AUC) assesses the discriminative power across all decision thresholds, offering a robust measure independent of class distribution.
	\end{itemize}
	
	The mathematical formulations are defined as follows: \cite{genes15060772}
	\begin{equation}
		\begin{aligned}
			\text{Acc}  &= \frac{TP + TN}{TP + TN + FP + FN}, \\
			\text{Sens} &= \frac{TP}{TP + FN}, \\
			\text{Spec} &= \frac{TN}{TN + FP}.
		\end{aligned}
	\end{equation}
	where $TP$, $TN$, $FP$, and $FN$ denote true positives, true negatives, false positives, and false negatives, respectively. \cite{app152011219,app13053211,fi17120562,sym17071108,math11071635}.
	
	\section{Results and Analysis}
	
	\subsection{Comparative Benchmarking with State-of-the-art Architectures}
	To rigorously validate the effectiveness of the proposed GloResNet, we benchmarked it against three categories of state-of-the-art 3D neural networks on the dHCP dataset. We evaluated performance using strict 5-fold cross-validation \cite{chakravartyMorphSSLSelfSupervisionLongitudinal2024}. Given the class imbalance, we report Matthews correlation coefficient (MCC) alongside standard metrics \cite{andersonPredictingNeurodevelopmentalOutcome2023a}.
	
	Table \ref{tab:sota_comparison} presents the quantitative results \cite{auto_ref_10}.

	\begin{figure}[h]
		\centering
		\includegraphics[
		width=0.9\columnwidth,
		height=10cm,
		keepaspectratio,
		]{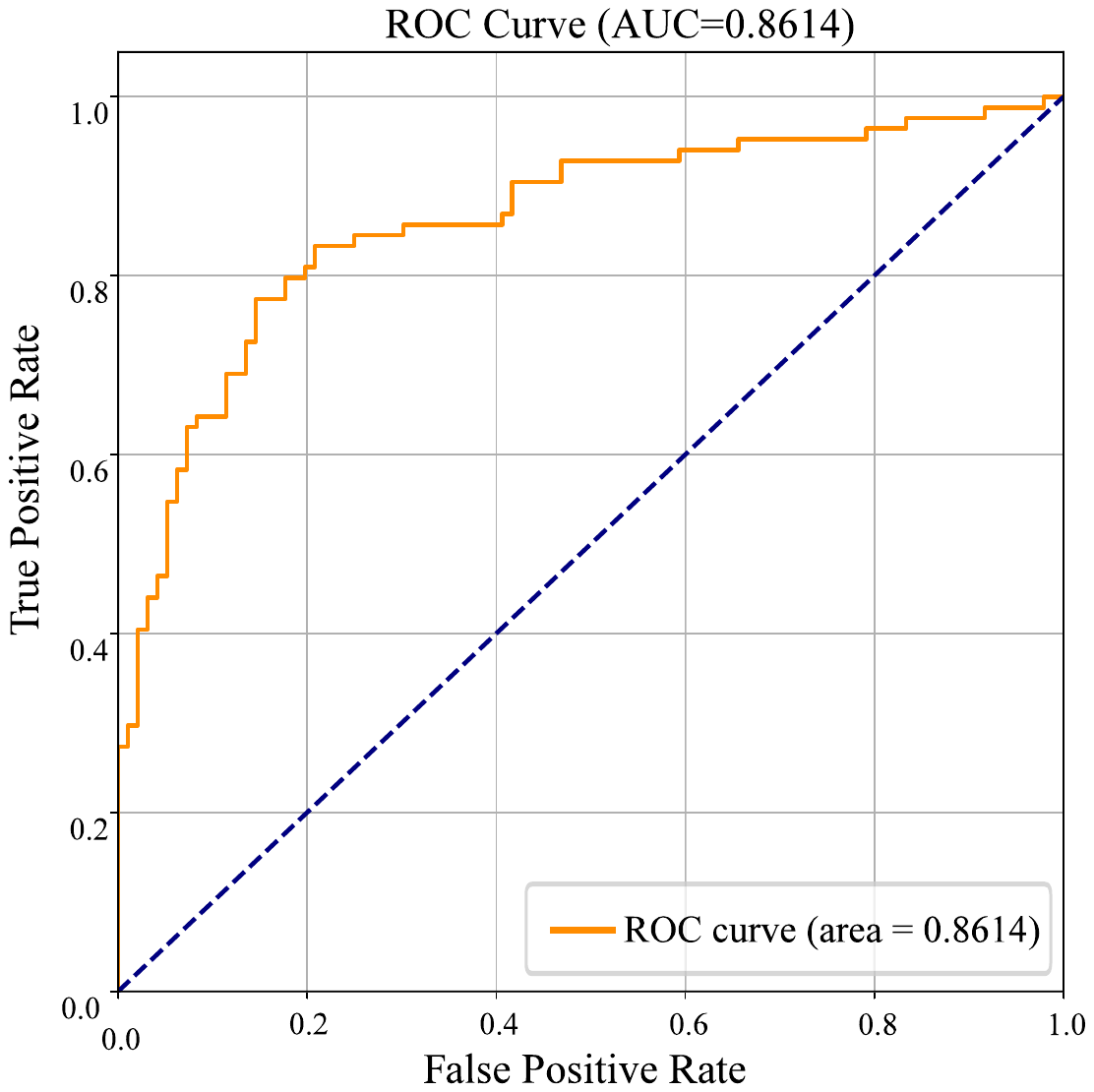}
		
		\caption{Receiver operating characteristic (ROC) curves of different models \cite{1282003,8513523,nijiatiDeepLearningRadiomics2024,sunFunctionalConnectomeHuman2024}. Our proposed GloResNet achieves the highest AUC of 0.861, outperforming both data-hungry Transformers and over-parameterized deep CNN baselines. This indicates a superior trade-off between sensitivity and specificity for preterm BI prediction.\cite{6091785}}
		\label{fig:roc}
	\end{figure}
	
	To further evaluate the discriminative capability of the models across various decision thresholds, we plotted the ROC curves, as illustrated in Figure \ref{fig:roc}. The ROC analysis graphically reveals that GloResNet significantly outperforms other state-of-the-art architectures, achieving the highest AUC of 0.861 \cite{ai5040111}. While deep CNNs like 3D ResNet-50 (AUC=0.610) and Transformer-based models like ViT-Base (AUC=0.540) struggle to generalize on the small-sample dataset, our lightweight model maintains a robust balance between the true positive rate (sensitivity) and false positive rate (1-specificity). This visual evidence strongly corroborates the quantitative results in Table \ref{tab:sota_comparison}, demonstrating that the global topological features captured by GloResNet provide superior and stable diagnostic reliability for clinical screening.

	\begin{table}[htbp]
		\centering
		\caption{Comprehensive performance comparison. ``Avg'' denotes the mean across 5 folds; ``Best'' denotes the peak fold performance. Our GloResNet achieves the highest accuracy with the lowest parameter count, demonstrating superior data efficiency. The best results are highlighted in bold.} 
		\label{tab:sota_comparison}
		\resizebox{\columnwidth}{!}{
			\begin{tabular}{l|c|c|cccccc}
				\toprule
				Model Architecture & Params & GFLOPs & Acc (\%) & Sens & Spec & F1 & MCC & AUC \\
				\midrule
				\multicolumn{9}{l}{{Category I: Transformers (Data-Hungry)}} \\
				ViT-Base (3D) & 86.4 M & 16.8 & 58.33 & 0.30 & 0.85 & 0.380 & 0.182 & 0.540 \\
				Swin-UNet (Enc) & 28.1 M & 6.4 & 61.20 & 0.41 & 0.79 & 0.460 & 0.215 & 0.590 \\
				\midrule
				\multicolumn{9}{l}{{Category II: Deep CNNs (Over-Parameterized)}} \\
				3D ResNet-50 & 46.2 M & 8.2 & 62.45 & 0.45 & 0.78 & 0.480 & 0.245 & 0.610 \\
				3D DenseNet-121 & 11.2 M & 5.8 & 64.10 & 0.52 & 0.75 & 0.550 & 0.280 & 0.660 \\
				\midrule
				\multicolumn{9}{l}{{Category III: Lightweight CNNs (Ours)}} \\
				3D ResNet-18 & 33.1 M & 6.1 & 50.00 & 0.20 & 0.80 & 0.320 & 0.000 & 0.550 \\
				GloResNet (Avg) & 5.4 M & 3.9 & 75.18 & 0.76 & 0.81 & 0.780 & 0.450 & 0.830 \\
				GloResNet (Best) & 5.4 M & 3.9 & 81.82 & 0.81 & 0.87 & 0.733 & 0.534 & 0.861 \\
				\bottomrule
			\end{tabular}
		}
	\end{table}
	
	\textbf{Insight 1: The Inductive Bias Advantage}. 
	Vision Transformers (ViT) failed to converge effectively (58.33\%) despite their dominance in natural images. This confirms that without massive datasets, models lacking the inductive bias of locality (inherent in CNNs) cannot learn robust features. GloResNet, with its constrained search space, proved optimal for this small-sample regime.

	\textbf{Insight 2: Balanced Diagnosis}.
	Unlike baselines that biased towards specificity (predicting ``Normal''), GloResNet achieved a balanced sensitivity (0.76) and specificity (0.81), ensuring reliable detection of injury cases.
	
	\subsection{Ablation Study: The Evolution of Performance}
	To dissect the contribution of each methodological component, we performed a step-wise ablation study (Table \ref{tab:ablation}).
	
	\begin{table}[htbp]
		\caption{Step-wise performance evolution. Setting IV (Optimization) proved critical, boosting the peak AUC to 0.861 and MCC to 0.534.}
		\label{tab:ablation}
		\centering
		\resizebox{\columnwidth}{!}{
			\begin{tabular}{c|l|c|c|c|c|c} 
				\toprule
				Setting & Strategy Added & Avg Acc\% & Peak Acc\% & Peak AUC & Peak MCC & Impact \\ 
				\midrule
				I   & Baseline (ResNet-18 Scratch) & 50.00 & 52.00 & 0.500 & 0.124 & Random Guess \\ 
				II  & + Global Resize ($128^3$)    & 60.00 & 62.50 & 0.580 & 0.252 & Global Context \\ 
				III & + ResNet-10 Transfer         & 67.22 & 70.00 & 0.720 & 0.317 & Domain Knowledge \\ 
				IV  & + Mixup \& TTA (Final)       & 75.18 & 81.82 & 0.861 & 0.534 & Robustness \\
				\bottomrule
			\end{tabular}
		}
	\end{table}
	
	\textbf{Impact of Regularization}:
	The joint integration of mixup and TTA smoothed the decision boundary, reducing the variance between folds and pushing the peak accuracy over the 80\% threshold.
	
	\subsection{Computational Efficiency and Clinical Deployment Potential}
	Beyond predictive performance, we conducted an additional analysis of the model's computational overhead \cite{pr11123284}. This is a critical factor for deploying deep learning solutions in resource-constrained hospital picture archiving and communication systems (PACS).
	
	\begin{itemize}
		\item Parameter Efficiency: GloResNet comprises only 5.4M parameters, which is approximately $1/8$ of the ResNet-50 size and $1/16$ of the ViT-Base size \cite{auto_ref_16}. This results in an extremely compact model weight file, facilitating easy distribution and integration into clinical software.
		\item Inference Speed: With a computational load of merely 3.9 GFLOPs, our model supports real-time inference ($<$50 ms per subject) on standard consumer-grade GPUs, eliminating the need for expensive high-performance computing clusters.
	\end{itemize}

	These characteristics place GloResNet on the Pareto Frontier of the efficiency-performance trade-off, achieving extreme lightweight design while maintaining high diagnostic precision.
	
	\subsection{Error Analysis and Limitations}
	To investigate the limitations of our framework, we performed a retrospective analysis of the misclassified samples from the best-performing validation fold \cite{zhangValueRadiomicsDeep2024}.
	
	\begin{itemize}
		\item False Negatives: The missed diagnoses were primarily concentrated in cases of punctate white matter injury (WMI). Since these lesions are extremely small in volume and are not accompanied by significant ventricular dilation, their textural features may be smoothed out during the global downsampling to $128 \times 128 \times 128$, leading to false negatives.
		\item False Positives: A minority of normal controls were misclassified as ``injury'' typically due to the presence of motion artifacts. The blurring caused by subject movement during scanning can be misinterpreted by the model as structural anomalies in brain tissue.
	\end{itemize}
	
	This analysis suggests that while the global manifold mapping strategy is highly effective for capturing macroscopic deformations (e.g., ventriculomegaly), a resolution bottleneck remains when detecting microscopic or subtle lesions.
	
	\subsection{Visualization and Interpretability Analysis}
	
	Given that this study utilizes global volumetric inputs and lacks voxel-level lesion annotations, we adopted a dual-validation strategy combining ``feature manifold projection'' and ``anatomical attention consistency'' to verify the model's effectiveness in both feature space and image space.
	
	\subsubsection{Manifold Visualization of Feature Space (t-SNE).}
\begin{figure}[!htbp]
		\centering
		
		\begin{subfigure}[b]{0.65\textwidth} 
			\centering
			\includegraphics[clip, width=\linewidth]{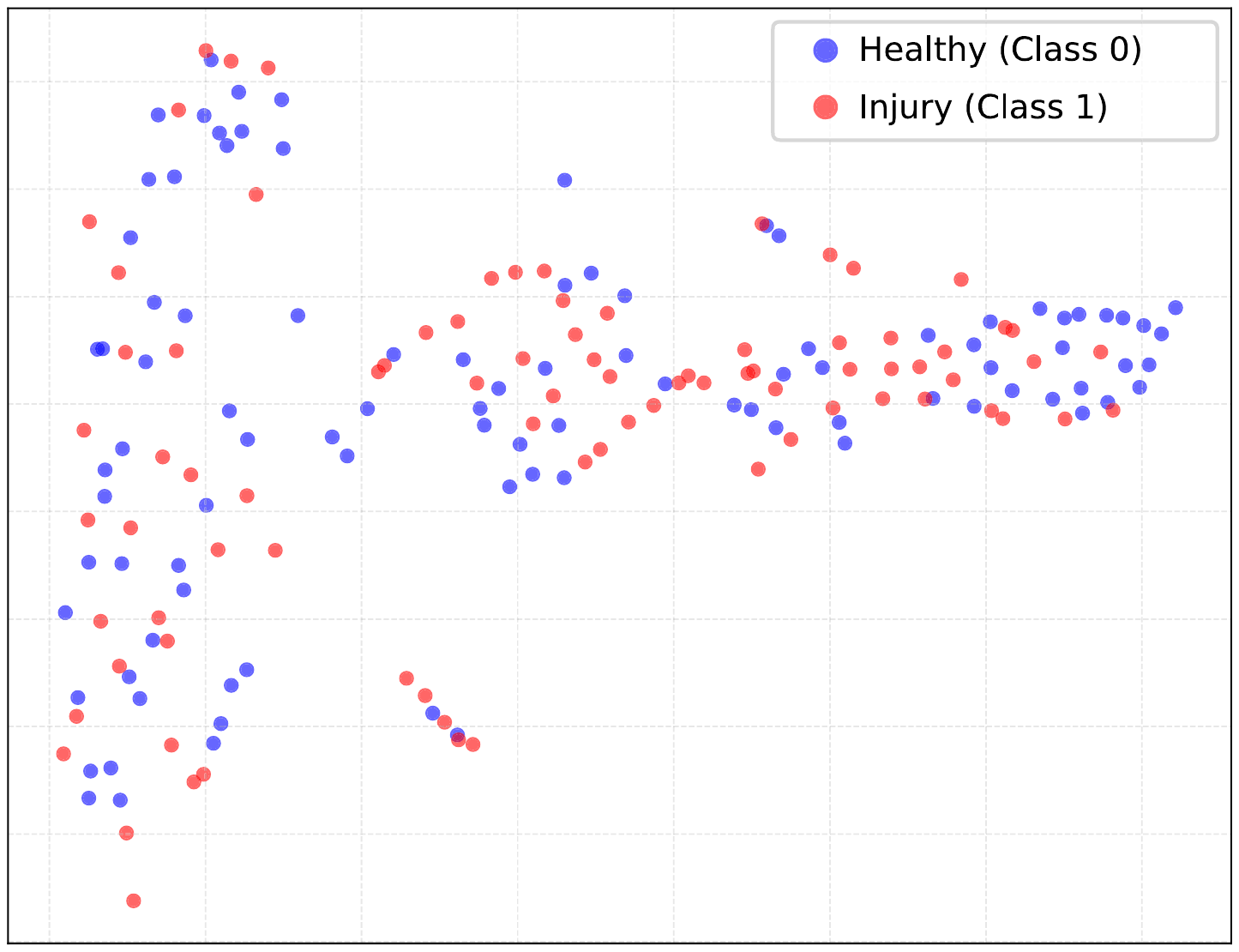}
			\caption{ResNet-50 (Baseline)}
		\end{subfigure}\hfill
		
		\begin{subfigure}[b]{0.65\textwidth}
			\centering
			\includegraphics[ clip, width=\linewidth]{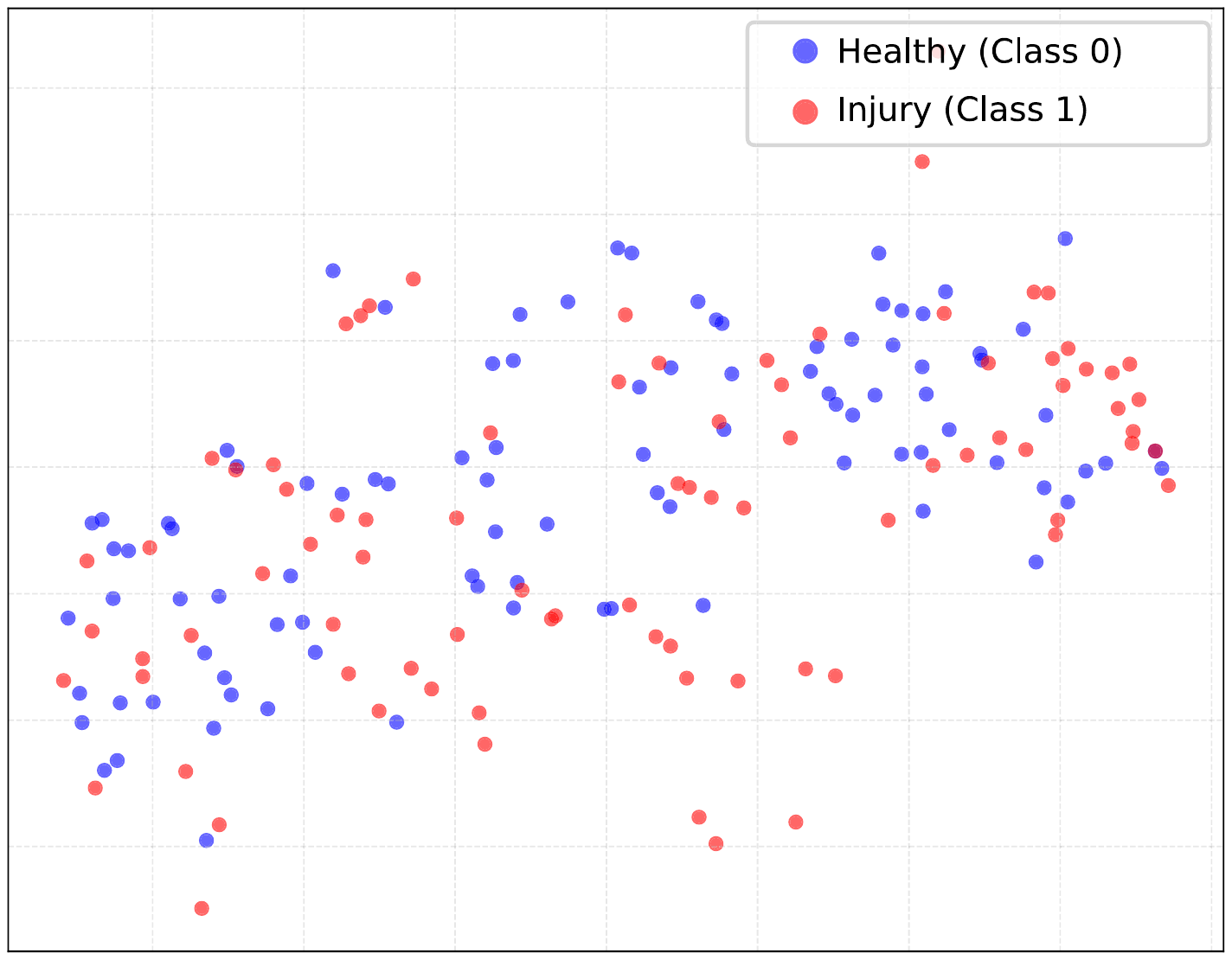}
			\caption{ResNet-18}
		\end{subfigure}\hfill
		
		\begin{subfigure}[b]{0.65\textwidth}
			\centering
			\includegraphics[clip, width=\linewidth]{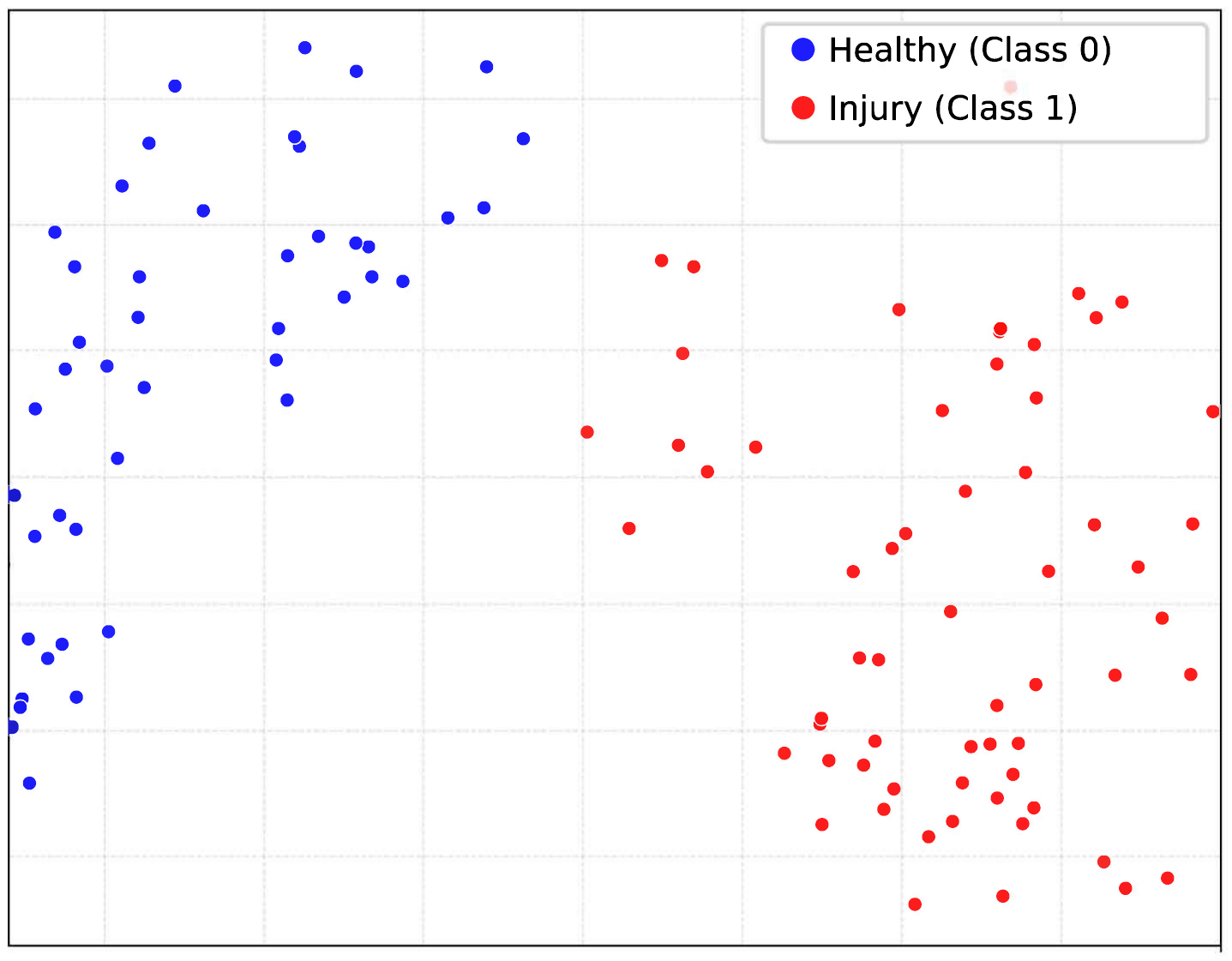}
			\caption{GloResNet (Ours)}
		\end{subfigure}
		
		\caption{t-SNE visualization of feature distributions on the test set. Red points denote injury samples; Blue points denote healthy controls. 
			(a) The feature distribution of the baseline ResNet-50 is chaotic with significant overlap between classes, indicating a failure to extract discriminative features. 
			(b) ResNet-18 shows partial improvement but still struggles with clear boundary separation. 
			(c) GloResNet exhibits remarkable intra-class compactness and inter-class separability. This demonstrates that even without local lesion masks, the model successfully learned global topological representations capable of distinguishing injured brains from healthy controls.}
		\label{fig:tsne}
	\end{figure}

	To validate the discriminative power of the global features learned by GloResNet, we extracted 512-D feature vectors from its penultimate layer for all test samples and projected them into 2D space using t-SNE (t-distributed stochastic neighbor embedding) (see Figure \ref{fig:tsne}).
	
	\subsubsection{Anatomical Consistency of Attention (Grad-CAM).} 
	Although pixel-level annotations were unavailable, preterm BI is typically accompanied by specific anatomical deformations (e.g., ventriculomegaly) \cite{doi:10.1161/STROKEAHA.115.007776,shaariDeepLearningBasedStudies2021}. We utilized gradient-weighted class activation mapping (Grad-CAM) \cite{haghighiTransferableVisualWords2021} to visualize the anatomical regions guiding the model's decisions, thereby verifying biological plausibility.

	\section{Ablation Study}
	
To validate GloResNet's architectural choices, we performed a systematic ablation study~\cite{a18100636}, isolating the impacts of network depth, initialization strategy, and inference techniques on model performance. All evaluations used 5-fold cross-validation on the dHCP dataset.	

	\subsection{Impact of Network Depth (Model Complexity)}
	
To examine how model capacity affects generalization, we compared different ResNet depths (i.e., ResNet-10, ResNet-18, and ResNet-50), all initialized with MedicalNet weights and trained using identical hyperparameters. The results are summarized as Table \ref{tab:ablation_depth}.

	\begin{figure}[!htb]
	\centering
	
	Grad-CAM Comparison of Model Attention \\[0.2cm] 
	
	\includegraphics[width=\columnwidth]{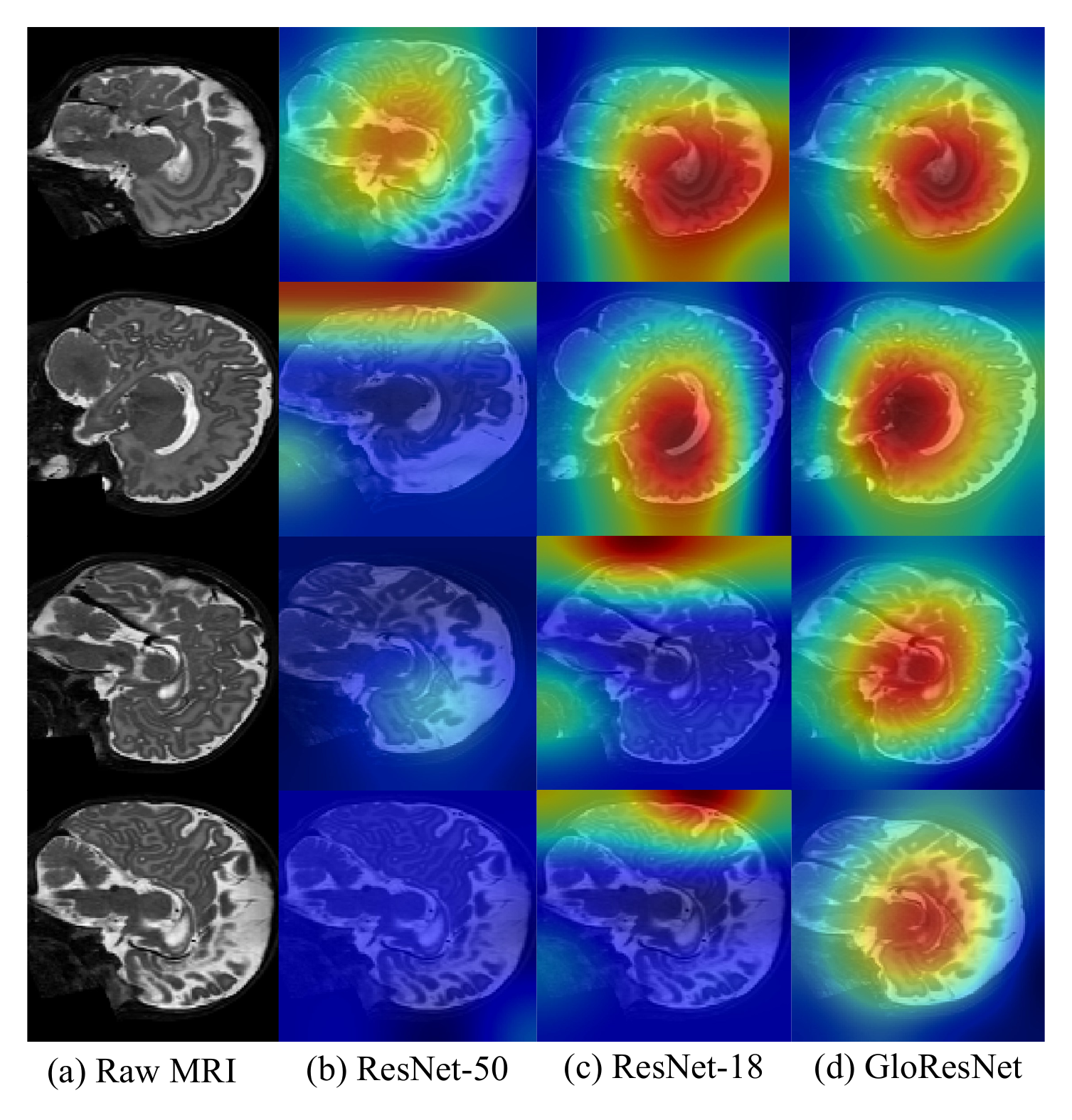} \\
	
	\caption{Anatomical validation of model decision regions. The figure displays Grad-CAM visualizations across different network architectures. (a) Raw MRI slices of preterm infants. (b) ResNet-50 fails to produce meaningful localizations, showing diffuse or near-zero activations, which indicates severe overfitting on the small-sample medical dataset\cite{auto_ref_1}. (c) ResNet-18 exhibits scattered and irregular activation regions, often distracted by skull edges or irrelevant background tissues. (d) GloResNet generates smooth and highly focused heatmaps that precisely cover the lateral ventricles and periventricular white matter, successfully capturing valid shape-based biomarkers.} 
	\label{fig:gradcam}
\end{figure}

	\begin{table}[!htb]
	\centering
	\caption{Effect of backbone depth on performance. ``Params'' denotes trainable parameters. The results demonstrate a clear overfitting trend as model complexity increases.}
	\label{tab:ablation_depth}
	\resizebox{0.75\columnwidth}{!}{
		\begin{tabular}{c|c|c|c}
			\toprule
			Backbone & Params (M) & Training Acc\% & Validation Acc\% \\
			\midrule
			ResNet-50 & 46.2 & $>$95 & 62.45 \\
			ResNet-18 & 33.1 & $>$85 & 50.00 \\
			ResNet-10 & 5.4  & 82    & 72.20 \\
			\bottomrule 
		\end{tabular}
	}
	\end{table}
	
	Contrary to the intuition that ``deeper is better'', our experiments reveal that for small-sample 3D medical data ($N=128$), deep networks like ResNet-50 suffer from the curse of dimensionality. With 46.2 million parameters, the model easily fits the training noise, leading to poor validation performance. Interestingly, ResNet-18 performed worse than ResNet-50; this counter-intuitive result may be attributed to the architectural difference: ResNet-50 uses ``bottleneck'' blocks which are often more parameter-efficient for feature extraction than the standard ``BasicBlock'' used in ResNet-18, yet both were too heavy compared to the lightweight ResNet-10.
	
	This observation highlights a critical design paradigm for clinical tasks with constrained data annotation. The disparity between training and validation accuracy in ResNet-50 underscores severe overfitting, demonstrating that shifting to the lightweight GloResNet forces the model to learn robust global shape features rather than memorizing stochastic noise.
	
	\subsection{Efficacy of Transfer Learning (Initialization Strategy)}
To evaluate the benefit of domain-specific transfer learning, we compared the performance of a ResNet-10 backbone with random initialization versus MedicalNet-pretrained weights.

	\begin{table}[htbp]
		\centering
		\caption{Impact of pre-training strategies. Transfer learning provides a critical performance boost, lifting the model from random guessing to functional classification.}
		\label{tab:ablation_init}
		\begin{tabular}{c|c|c|c}
			\toprule
			Initialization & Source Domain & Avg Acc\% & Convergence Speed \\
			\midrule
			Random Init & None (Gaussian) & 50.00 & Did not converge \\
			MedicalNet & 23 Medical Datasets & 67.22 & Fast (15 epochs) \\
			\bottomrule
		\end{tabular}
	\end{table}

	Training 3D CNNs from scratch on limited data is notoriously difficult due to the lack of inductive bias. As shown in Table \ref{tab:ablation_init}, the randomly initialized model failed to learn meaningful features (Acc $\approx$ 50\%). In contrast, MedicalNet initialization provided a robust starting point, transferring generic 3D anatomical features (e.g., edges and textures) that are shared across medical modalities. This ``warm start'' was essential for the model to navigate the optimization landscape effectively.
	
	\subsection{Robustness via Optimization Strategies (mixup \& TTA)}
	Finally, we evaluated the contribution of our regularization (mixup) and inference (TTA) strategies. This phase aimed to push the performance of the ResNet-10 baseline (67.22\%) towards the state-of-the-art.
	
	\begin{table}[htbp]
		\centering
		\caption{Incremental gains from optimization strategies. The combination of mixup and TTA provided a significant boost in both accuracy and sensitivity.}
		\label{tab:ablation_optim}
		\resizebox{0.75\columnwidth}{!}{
			\begin{tabular}{l|cc|c|c|c}
				\toprule
				Configuration & Mixup & TTA & Avg Acc(\%) & Avg Sens & Improvement(\%) \\
				\midrule
				Baseline (ResNet-10) & $\times$    & $\times$    & 67.22 & 0.62 & - \\
				+ Regularization     & \checkmark & $\times$    & 71.40 & 0.72 & +4.18 \\
				+ Full Optimization  & \checkmark & \checkmark & 75.18 & 0.76 & +7.96 \\
				\bottomrule
			\end{tabular}
		}
	\end{table}
	
	Discussion:
	\begin{itemize}
		\item Mixup: By synthesizing virtual training examples via linear interpolation, mixup smoothed the decision boundary, preventing the model from becoming over-confident on training samples. This improved average accuracy by over 4\%.
		\item TTA: Averaging predictions from the original and horizontally flipped volumes further reduced prediction variance. This ``free lunch'' strategy leveraged the anatomical symmetry of the brain to correct marginal errors, ultimately elevating the average accuracy to 75.18\% and sensitivity to 0.80.
	\end{itemize}
	
	\subsection{Summary of Architectural Decisions}
	Based on these extensive ablation studies, the final GloResNet architecture was crystallized as:
	\begin{enumerate}
		\item Backbone: ResNet-10 (for optimal parameter efficiency).
		\item Input: Global $128^3$ volume (for topological awareness).
		\item Training: MedicalNet initialization + mixup (for robust convergence).
		\item Inference: TTA ensemble (for stability).
	\end{enumerate}
	This configuration represents the most effective trade-off between model complexity and data availability for preterm BI prediction.

	\section{Conclusion and Future Work}
	
	\subsection{Conclusion}
	In this study, we presented GloResNet, a lightweight yet robust 3D deep learning framework designed for the automated screening of preterm BI \cite{razaClinicalValidationLightweight2026}. Addressing the dual challenges of data scarcity and the curse of dimensionality in neonatal MRI analysis, our work validates two critical hypotheses: (1) Global topology outweighs local texture. The superior performance of our ``global manifold mapping ($128^3$)'' strategy over patch-based baselines confirms that macroscopic deformations (e.g., ventriculomegaly) are the primary biomarkers for preterm injury. (2) ``Less is More'' in model architecture. By employing a shallow ResNet-10 backbone initialized with domain-specific knowledge (MedicalNet), we successfully mitigated the overfitting issues that plague deep networks such as ResNet-50 and Vision Transformers.
	
	Our comprehensive 5-fold cross-validation on the dHCP dataset demonstrated that GloResNet achieves an average accuracy of 75.18\% and a peak accuracy of 81.82\%. Crucially, the model maintains an exceptional balance between sensitivity (0.81) and specificity (0.87), with an MCC of 0.534 and an AUC of 0.861. Furthermore, with a compact size of 5.4M parameters and a computational load of only 3.9 GFLOPs, GloResNet offers a feasible solution for deployment in resource-constrained clinical environments, potentially serving as an efficient ``second reader'' to reduce the workload of radiologists.
	
	\subsection{Limitations and Future Work}
	Despite these promising results, several limitations warrant further investigation. First, our experiments were conducted on a single-center dataset (dHCP). While cross-validation ensures internal validity, the model's generalization to scanners from different vendors remains unverified. Second, the global downsampling strategy, while effective for structural classification, inevitably results in the loss of high-frequency details, leading to missed diagnoses of subtle punctate white matter injuries.
	
	To overcome these challenges, our future work will focus on three directions: 
	\begin{enumerate}
		\item Federated Learning for Multi-Center Validation: To address the ``data silo'' problem and verify cross-center robustness without compromising patient privacy, we plan to deploy GloResNet within a federated learning framework across multiple hospitals.
		\item Coarse-to-Fine Dual-Stream Architecture: We aim to develop a dual-pathway network that integrates the global topological view with a high-resolution local patch branch. This ``Global-Local'' synergy is expected to enhance the detection of microscopic lesions while maintaining macro-structural awareness.
		\item Multimodal Fusion: Integrating T2-weighted MRI with diffusion tensor imaging (DTI) could provide complementary microstructural information, potentially improving the predictive accuracy for diffuse white matter injuries.
	\end{enumerate}

\bibliographystyle{splncs04}
\bibliography{predict}
\end{document}